\documentclass[journal]{IEEEtran}

\usepackage[numbers,sort&compress]{natbib}
\usepackage{amsmath,amssymb,amsthm}
\usepackage{graphicx}
\usepackage{balance}
\usepackage[table]{xcolor}
\usepackage{array}
\usepackage{multirow}
\usepackage{booktabs}
\usepackage{dblfloatfix}
\usepackage{url}
\newtheorem{theorem}{Theorem}

\newtheorem{assumption}{Assumption}
\theoremstyle{remark}

\newcommand{\contribitem}[2]{%
  \par\noindent
  \hangindent=1.25em
  \hangafter=1
  \makebox[1.25em][l]{#1)}#2%
}

\begin{document}

\title{CARM: Conformal Adaptive Risk Masking for Reliable Scientific Super-Resolution}

\author{
Guangtao Zhang\textsuperscript{1},
Haonan Wu\textsuperscript{1},
Junxuan Gao\textsuperscript{1},
Zhengyang Liu\textsuperscript{1},\\
Zhaoxi Zeng\textsuperscript{1},
Zhihao Li\textsuperscript{4},
Jianguo Huang\textsuperscript{3},
and Zhiqiang Liu\textsuperscript{2}%
\thanks{This work was supported in part by the National Natural
Science Foundation of China under Grant 42450181 and in part by
the Research Grants Council of the Hong Kong Special Administrative
Region, China, under Grant AoE/P-601/23-N.
(Corresponding authors: Zhiqiang Liu; Jianguo Huang.)}
\thanks{\textsuperscript{1}SandGold AI Research, China;
\textsuperscript{2}Southern University of Science and Technology,
China (e-mail: liuzq@sustech.edu.cn);
\textsuperscript{3}Nanyang Technological University, Singapore
(e-mail: jianguo.huang@ntu.edu.sg);
\textsuperscript{4}The Hong Kong University of Science and
Technology (Guangzhou), China.}
}

\maketitle

\begin{abstract}
Deep learning-based super-resolution (SR) can reconstruct fine-scale
Earth-observation fields from coarse observations, but reconstruction
accuracy alone does not provide a calibrated rule for identifying
reliable spatial regions. Conformal Mask Calibration (CMC), built on
conformal risk control, converts a spatial uncertainty score into a
trust mask and controls the expected worst-case local distortion under
exchangeability. However, its fixed threshold can become miscalibrated
in Earth-observation streams affected by seasonality and episodic
events. We propose \textit{CARM} (Conformal Adaptive Risk Masking), a
statistical framework for adaptive spatial risk control under temporal
distribution shift. CARM maintains experts with heterogeneous update
rates and aggregates their threshold proposals online using an
asymmetric surrogate loss, balancing rapid adaptation to distribution
changes against stability during relatively stationary periods. All
updates are causal and use only mature feedback available after a
fixed delay, allowing CARM to operate with any SR model that provides
a spatial uncertainty score. We prove that the time-averaged
mask-conditioned risk converges to at most $\alpha$ at an $O(1/T)$
rate, without distributional assumptions and for any choice of
aggregation weights across experts. On a daily South China Sea
chlorophyll-\textit{a} SR stream, CARM reduces the empirical violation
rate from \(43.49\%\) to \(9.38\%\) and the moving-average exceedance
frequency from \(47.29\%\) to \(1.22\%\) at \(\alpha=0.10\), while
retaining a non-trivial trusted region. Experiments on ERA5 wind and
OSTIA sea-surface temperature further show consistent relative
improvements across multiple SR backbones.
\end{abstract}


\section{Introduction}

High-resolution satellite observations of ocean fields are important
for studying marine ecosystem dynamics, primary productivity, and
the global carbon
cycle~\citep{mcclain2009decade,field1998primary}.
However, satellite ocean-color sensing faces an inherent
spatiotemporal trade-off: frequent observations are generally
available at relatively coarse spatial resolutions, whereas
finer-resolution sensors revisit a given location less often.
Deep learning-based super-resolution (SR) provides a practical means
of reconstructing fine-scale ocean fields from routinely available
coarse satellite
observations~\citep{lloyd2021optically,liu2021super,yang2025super}.
Yet reconstruction accuracy alone does not indicate which spatial
regions are sufficiently reliable for downstream scientific
analysis.

To quantify the reliability of predictions, existing uncertainty quantification methods provide pixel-wise uncertainty scores, i.e., predictive variability, in image SR~\citep{saranathan2023perpixel,Liu_2023_CVPR}. Such scores, however, remain purely descriptive: they offer no spatial trust mask controlled against an explicit fidelity-risk target.
Conformal prediction provides distribution-free uncertainty quantification under exchangeability, while Conformal Risk Control (CRC) extends conformal guarantees to user-defined risk functionals~\citep{angelopoulos2022gentleintroductionconformalprediction,angelopoulos2024conformal}. Building on the CRC framework, Conformal Mask Calibration (CMC)~\citep{adame2025image} converts a spatial uncertainty score into a risk-controlled trust mask: it calibrates a threshold once on a calibration dataset and, at deployment, marks as trusted every pixel whose uncertainty score falls below this threshold. This procedure inherits the CRC guarantee that the expected fidelity risk over the trust mask does not exceed a user-specified level $\alpha$, provided the calibration and deployment data remain exchangeable.
In Earth-observation streams, however, seasonality and episodic events can alter the data distribution~\citep{singh2024uncertainty}, breaking exchangeability and voiding the CRC guarantee for the calibrated threshold.

To address this limitation, adaptive conformal methods update calibration thresholds online under distribution
shift~\citep{gibbs2021adaptive,angelopoulos2023conformal_pid,
marx2025calibrated}. However, existing approaches have been developed primarily for scalar prediction targets rather than the dense spatial trust masks required in our setting. Spatial masking must additionally accommodate delayed error feedback and adapt across changing temporal scales. Moreover, a fixed update step size introduces two further problems: a value small enough to avoid overreacting to transient fluctuations is also slow to relax once conditions improve, leaving the trust mask unnecessarily conservative; and a value tuned to one dataset's drift speed and noise level typically requires re-tuning on another.

We close these gaps with CARM (Conformal Adaptive Risk Masking), which maintains multiple experts with different update rates and aggregates their threshold proposals online. CARM uses only mature feedback available after a fixed delay and applies an asymmetric loss that penalizes risk violations more heavily than unnecessary conservatism. CARM produces causal thresholds, and its time-averaged mask-conditioned risk provably converges to at most $\alpha$ at an $O(1/T)$ rate, without distributional assumptions and for any choice of aggregation weights across experts. Experiments on South China Sea Chl-a, ERA5 wind, and OSTIA sea surface temperature show consistent relative improvements over static CMC across multiple Earth-observation variables, target risk levels, and SR backbones.

The main contributions are summarized as follows:
\contribitem{1}{We identify the deployment mismatch of static conformal trust masks under temporal distribution shift.}
\contribitem{2}{We propose CARM, a post-hoc framework that combines multi-rate experts, causal delayed-feedback updates, and asymmetric online aggregation. Moreover, we prove that CARM asymptotically controls the time-averaged mask-conditioned risk at the target level without distributional assumptions.}
\contribitem{3}{We validate CARM on South China Sea Chl-a, ERA5 wind, and OSTIA sea surface temperature using multiple SR backbones, showing consistent improvements over static conformal mask calibration.}

\section{Problem Setup and Motivation}
\label{subsec:problem_motivation}

We consider online trusted-region prediction for super-resolution
under temporal distribution shift. On day $t$, a predictor maps a
low-resolution field $X_t\in\mathbb{R}^{H\times W}$ to a
super-resolved field
$\hat{Y}_t\in\mathbb{R}^{sH\times sW}$, where $H$ and $W$ are the
low-resolution spatial dimensions and $s$ is the upscaling factor.
The corresponding high-resolution reference is
$Y_t\in\mathbb{R}^{sH\times sW}$. Let
$U_t\in\mathbb{R}^{sH\times sW}$ denote the uncertainty map of the
prediction. A trust mask under threshold $\tau$ is
\begin{equation}
\label{eq:trust_mask}
M_t(\tau)
=
\left\{
p\in\Omega_t:U_t(p)\leq\tau
\right\},
\end{equation}
where $p$ indexes a high-resolution pixel and $\Omega_t$ denotes the
set of valid pixels on the high-resolution grid. The trusted-area
ratio under threshold $\tau$ is
$c_t(\tau)=|M_t(\tau)|/|\Omega_t|$. We define the bounded local
distortion as
\[
\tilde e_t(p)
=
\min\left\{
1,\,
\frac{|\hat Y_t(p)-Y_t(p)|}{b}
\right\}
\in[0,1],
\]
where $b>0$ is a reference scale, and the mask-conditioned risk as
the largest distortion within the trusted area:
\begin{equation}
r_t(\tau)
=
\max_{p\in M_t(\tau)}
\tilde e_t(p).
\label{eq:risk}
\end{equation}
Unlike a spatially averaged error, this risk prevents a small number
of large local distortions from being hidden by accurate predictions
elsewhere in the retained region. It therefore evaluates whether
every retained location remains within the intended fidelity level.
Increasing $\tau$ retains a larger spatial region but may also include
locations with greater reconstruction error, thereby creating a
reliability--coverage trade-off.

Given this risk, we first calibrate a static threshold on a
calibration set via \textit{Conformal Mask Calibration (CMC)}:
\begin{equation}
\label{eq:cmc_threshold}
\hat{\tau}_{\alpha}^{\mathrm{CMC}}
=
\sup\left\{
\tau:
\frac{\sum_{i=1}^{n}r_i(\tau)+1}{n+1}
\leq\alpha
\right\},
\end{equation}
where $n$ is the number of calibration samples and $\alpha$ is the
target risk level. If the calibration samples together with the test
sample are exchangeable, this rule guarantees
\begin{equation}
\mathbb{E}\big[
r_{n+1}(\hat{\tau}_{\alpha}^{\mathrm{CMC}})
\big]
\leq\alpha
\end{equation}
for the $(n+1)$-th example
\cite{angelopoulos2022gentleintroductionconformalprediction,
angelopoulos2024conformal}.
This static threshold provides the non-adaptive deployment baseline,
while calibration-set uncertainty scores are also used to construct
the candidate grid $\mathcal G$ employed by the adaptive procedure
introduced in Section~III.

However, real-world ocean streams are not identically distributed:
seasonality, regime shifts, and episodic events such as mesoscale
eddies and marine heatwaves alter the relationship between uncertainty
and SR reconstruction error~\citep{singh2024uncertainty}.
Consequently, a threshold calibrated once on historical data can
become unsafe under unfamiliar regimes or overly conservative under
benign conditions. In the former case, the static threshold retains
regions whose realized errors are no longer compatible with the
target risk; in the latter, it rejects regions that could still be
used reliably.

A direct remedy is online recalibration using matured feedback: the
high-resolution reference for day $t$ becomes available only after a
delay of $d$ days, so on day $t$ we can only use decisions whose
outcomes have already matured. Let
\begin{equation}
\mathcal B_t
=
\{t-d-w+1,\ldots,t-d\}
\end{equation}
denote the $w$ most recent matured decisions available on day $t$,
where $w$ is the rolling-window length. A single-rate update with
fixed step size $\eta$ can be summarized as
\begin{equation}
\begin{gathered}
\bar v_t
=
\frac{1}{w}
\sum_{j\in\mathcal B_t}
\mathbf{1}\{r_j(\tau_j)>\alpha\},\\
\alpha^e_{t+1}
=
\alpha^e_t+\eta(\alpha-\bar v_t),
\end{gathered}
\label{eq:single_rate_update}
\end{equation}
where $\tau_j$ is the threshold actually deployed on day $j$, and
$\alpha_t^e$ is the internal target used for online recalibration.
When the recent violation frequency exceeds the target, the update
decreases the internal level and makes subsequent threshold selection
more conservative. Conversely, when violations remain below the
target, the internal level is relaxed to recover usable trusted area.

\begin{figure}[t]
    \centering
    \includegraphics[width=\columnwidth]
    {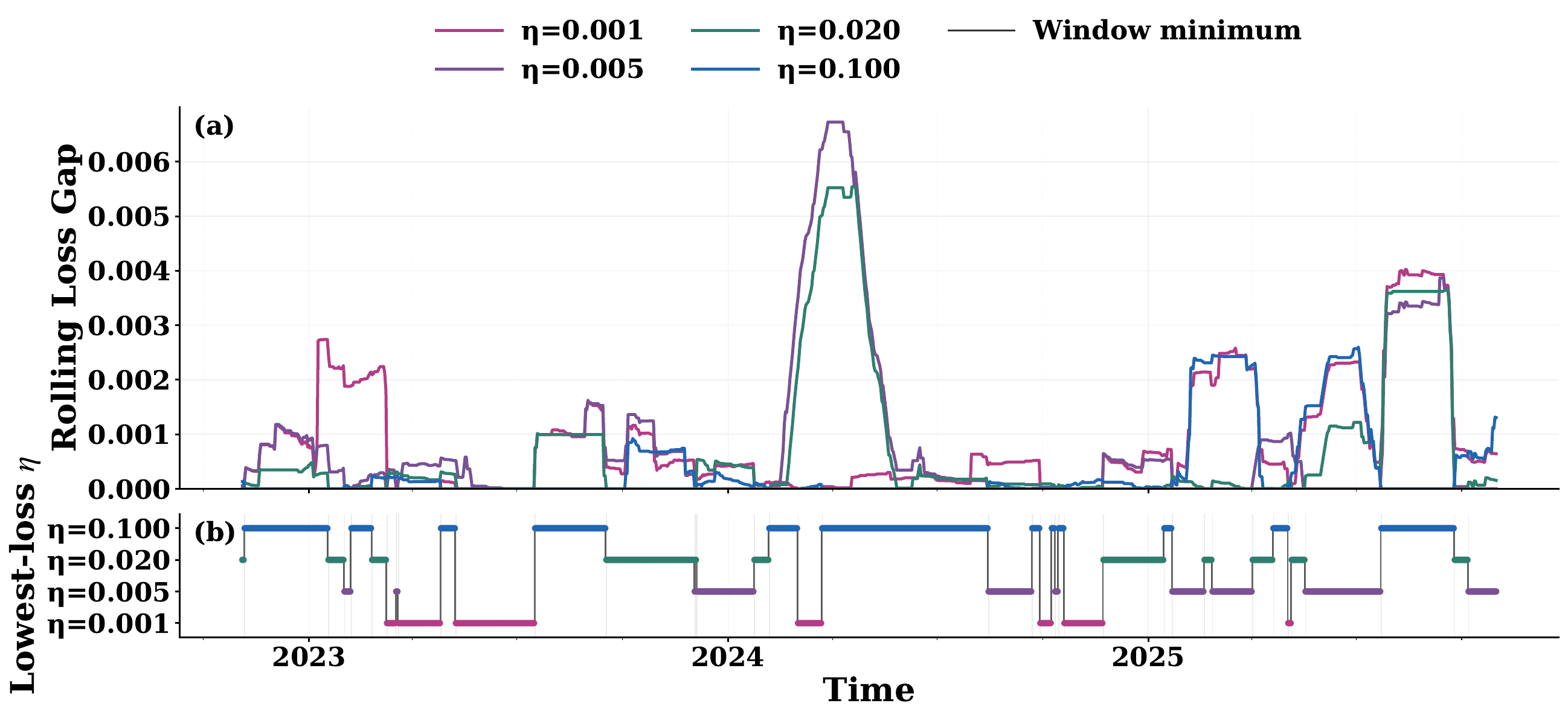}
    \caption{
    Fixed-step sensitivity on the South China Sea stream at
    $\alpha=0.15$ with $w=60$.
    Top: 60-day rolling-loss gap of each fixed-step expert relative
    to the minimum rolling loss among the experts in each window.
    Bottom: the adaptation rate $\eta$ attaining the minimum rolling
    loss in the corresponding window.
    }
    \label{fig:motivation_eta_shift}
\end{figure}

In practice, large $\eta$ values respond rapidly to persistent
changes but may overreact to transient fluctuations, whereas small
values are stable but adapt slowly.
Figure~\ref{fig:motivation_eta_shift} shows that the adaptation rate
attaining the lowest rolling loss changes repeatedly over time,
suggesting that no single fixed rate is uniformly favored throughout
deployment.

These observations motivate our method, \textit{CARM} (Conformal
Adaptive Risk Masking), which maintains experts with different step
sizes and aggregates their threshold proposals according to matured
performance rather than committing to a single fixed threshold or
adaptation rate.

\begin{figure*}[t]
    \centering
    \includegraphics[width=0.94\textwidth]{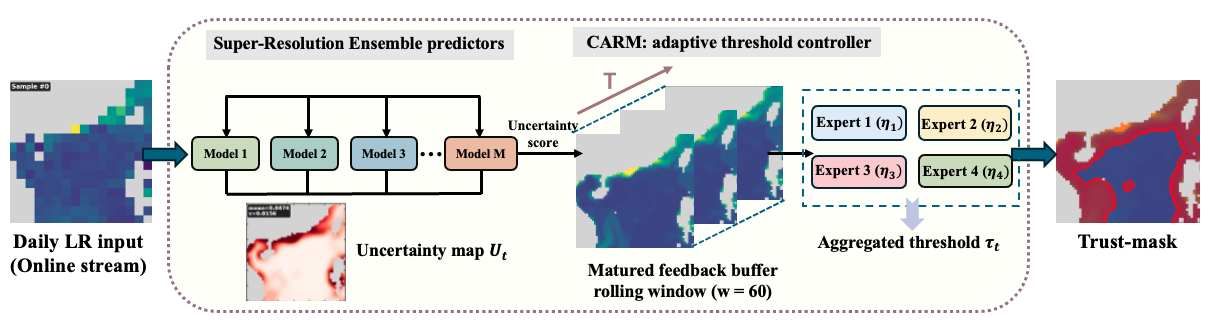}
    \caption{Overview of CARM. Experts with different step sizes update their threshold proposals. The
    proposals are aggregated into the deployed threshold $\tau_t$, which is applied to $U_t$ to generate the trust mask.}
    \label{fig:framework}
\end{figure*}

\section{Our method}
\label{subsec:carm}

CARM is a post-hoc framework that computes a deployment threshold $\tau_t$ using only feedback matured by day $t$, without retraining the underlying super-resolution predictor. Applying $\tau_t$ to the current uncertainty map $U_t$ via Eq.~\eqref{eq:trust_mask} yields the trust mask $M_t(\tau_t)$. Concretely, CARM lets $K$ experts, indexed by $k\in\{1,\ldots,K\}$ and each running at a distinct step size $\eta_k$, propose thresholds independently. It then combines these proposals into $\tau_t$ by upweighting whichever experts have kept their realized risk closest to $\alpha$ on recently matured decisions.

Each expert $k$ follows the same recalibration rule as
Eq.~\eqref{eq:single_rate_update}: it maintains a buffer
$\mathcal B_t^{(k)}=\{t-d-w+1,\ldots,t-d\}$ of its $w$ most recent
decisions whose outcomes have matured after the $d$-day feedback
delay, an empirical violation frequency $\bar v_t^{(k)}$ over this
buffer, and an internal target level $\alpha_t^{e,(k)}$, updated as
\begin{equation}
\label{eq:expert_adaptation}
\begin{gathered}
\bar v_t^{(k)}
=
\frac{1}{w}
\sum_{j\in\mathcal B_t^{(k)}}
\mathbf{1}
\left\{
r_j\!\left(\tau_j^{(k)}\right)>\alpha
\right\},\\
\alpha_{t+1}^{e,(k)}
=
\alpha_{t}^{e,(k)}
+
\eta_k
\left(
\alpha-\bar v_{t}^{(k)}
\right).
\end{gathered}
\end{equation}
A violation rate above $\alpha$  reduces the expert's
internal target, making subsequent threshold selection more conservative.

Candidate thresholds are drawn from the fixed grid $\mathcal{G}$,
constructed from calibration-set uncertainty percentiles. Let
$\mathcal{G}_t\subseteq\mathcal{G}$ denote the candidate set used at
deployment step $t$. For each $\tau\in\mathcal{G}_t$, let
$\hat q_t(\tau)$ denote its empirical violation frequency over the
matured candidate records. Expert $k$ selects
\begin{equation}
\label{eq:expert_threshold_selection}
\tau_t^{(k)}
\in
\arg\min_{\tau\in\mathcal{G}_t}
\left|
\hat q_t(\tau)-\alpha_t^{e,(k)}
\right|
\end{equation}
with ties resolved in favor of the larger threshold. Thus,
$\hat q_t(\tau)$ evaluates candidate thresholds, whereas
$\bar v_t^{(k)}$ evaluates the realized decisions of expert $k$ and
updates its internal target.

The experts' proposals are combined into the deployed threshold
using the current weights $\omega_t^{(k)}$, carried over from the
previous day:
\begin{equation}
\label{eq:deployment}
\tau_t = \sum_{k=1}^{K} \omega_t^{(k)} \tau_t^{(k)}.
\end{equation}

By day $t$, the reference $Y_{t-d}$ has also just arrived after the
$d$-day delay, making the realized risk $r_{t-d}(\tau_{t-d}^{(k)})$
of the decision expert $k$ made on day $t-d$. Expert $k$ incurs the loss
\begin{equation}
\label{eq:expert_loss}
\ell_t^{(k)}
=
\begin{cases}
\left[
\alpha-r_{t-d}\!\left(\tau_{t-d}^{(k)}\right)
\right]^2,
&
r_{t-d}\!\left(\tau_{t-d}^{(k)}\right)\leq\alpha,\\[3pt]
r_{t-d}\!\left(\tau_{t-d}^{(k)}\right)-\alpha,
&
r_{t-d}\!\left(\tau_{t-d}^{(k)}\right)>\alpha.
\end{cases}
\end{equation}
The loss mildly penalizes conservativeness below the target and more
strongly penalizes target violations, based on this single freshly
matured day rather than the $w$-day average used for
$\bar v_t^{(k)}$. This loss updates each weight for tomorrow's deployment:
\begin{equation}
\label{eq:weight_update}
\omega_{t+1}^{(k)}
=
\frac{
\omega_t^{(k)}
\exp\!\left(-\ell_t^{(k)}\right)
}{
\sum_{j=1}^{K}
\omega_t^{(j)}
\exp\!\left(-\ell_t^{(j)}\right)
},
\end{equation}
where the expert
weights are initialized uniformly as $\omega_1^{(k)}=1/K$ for all
$k\in\{1,\ldots,K\}$. The overall procedure is illustrated in
Fig.~\ref{fig:framework}.
Each expert's internal target $\alpha_t^{e,(k)}$ is calibrated
against the user-facing level $\alpha$, as in
Eq.~\eqref{eq:expert_adaptation}. Next, we prove the asymptotic
risk-control guarantee. Let $\tau_{\min}:=\min\{\mathcal G\}$ denote
the smallest candidate threshold, so that $r_t(\tau_{\min})$ is the
lowest risk attainable on day $t$.

\begin{theorem}[Asymptotic average risk control]
\label{thm:main}
Let $\alpha' = 1 - \sqrt{1-\alpha}$. Suppose
$r_t(\tau_{\min})\le\alpha'$ at every deployment day. Then
CARM controls the average mask-conditioned risk at the user-facing
level:
\begin{equation}
\label{eq:main}
\limsup_{T\to\infty}\;\frac{1}{T}\sum_{t=1}^{T}r_t(\tau_t)\;\le\;\alpha ,
\end{equation}
with $O(1/T)$ convergence.
\end{theorem}
The guarantee does not require i.i.d.\ sampling or exchangeability
and holds for any aggregation weights. The proof and assumptions are
given in the supplementary material.

\section{Experiments}
\label{sec:experiments}

\subsection{Problem Setting}
\label{subsec:problem_setting}

\textbf{Setup and evaluation protocol.}
We evaluate CARM in a sequential scientific super-resolution setting
with chronologically ordered training, calibration, and test splits.
For each dataset--backbone combination, model parameters are optimized
using the training split, and the selected SR model remains fixed
throughout online deployment. For the FNO experiments, the calibration
split is also used for checkpoint selection, as detailed below. The
same split subsequently initializes the uncertainty-related quantities,
candidate threshold set, and online buffers used by CMC and CARM.

During testing, predictions and threshold updates are performed
sequentially day by day. The corresponding high-resolution reference
for day $t$ becomes available only after a fixed delay of $d=7$ days;
therefore, all adaptive updates use only matured feedback. Unless
otherwise stated, CARM uses a matured window of $w=60$ and target
levels $\alpha\in\{0.05,0.10,0.15\}$.

\textbf{Datasets and models.}
We evaluate CARM on three scientific datasets spanning distinct
geophysical variables: the South China Sea chlorophyll-\textit{a}
(Chl-\textit{a}) stream~\cite{copernicusmarine2018globalbiogeochemistry},
ERA5 wind fields~\cite{hersbach2023era5pressure}, and OSTIA sea-surface
temperature (SST) fields~\cite{copernicusmarine2024ostia}. We consider seven SR backbones:
FNO~\cite{li2021fno}, MWT~\cite{gupta2021mwt},
SRNO~\cite{wei2023srno}, HiNOTE~\cite{luo2024hinote},
SwinIR~\cite{liang2021swinir}, EDSR~\cite{lim2017edsr},
and ResShift~\cite{yue2023resshift}.

\textbf{Backbone training.}
The detailed South China Sea results and all ablation experiments use
FNO as the SR backbone. Each FNO ensemble member is trained
independently with AdamW using an initial learning rate of
$5\times10^{-4}$, weight decay of $10^{-3}$, and batch size $16$ for
up to $500$ epochs. A ReduceLROnPlateau scheduler monitors the MSE on
the calibration split with a reduction factor of $0.5$, patience of
$10$, and a minimum learning rate of $10^{-6}$. Early stopping is
applied with patience $40$, and the checkpoint attaining the lowest
calibration MSE is retained for subsequent uncertainty estimation and
sequential evaluation.

\textbf{Uncertainty estimation and CARM configuration.}
The main experiments use an ensemble of $M=5$ independently trained
SR models with different random seeds. Let
$\{\hat Y_t^{(m)}\}_{m=1}^{M}$ denote the ensemble predictions on day
$t$. For each HR pixel location, predictions from all ensemble
members within its $3\times3$ spatial neighborhood are collected to
compute a local joint variance. Its square root is used as the local
ensemble spread. The resulting spread map is further processed using
mask-aware Gaussian smoothing with $\sigma=1.5$, so that land pixels
do not contaminate nearby valid ocean locations. The resulting map
defines the predictive uncertainty map $U_t$.

The candidate threshold set $\mathcal G$ is constructed once from
the uncertainty distribution over the calibration split and remains
fixed throughout online deployment. Specifically, we use $L=201$
candidate thresholds corresponding to uniformly spaced percentile
levels from $0$ to $100$. The expert pool contains four adaptation
rates
$\eta\in\{0.001,0.005,0.02,0.1\}$.
The online buffers are warm-started from calibration records and
updated using only matured feedback.

\textbf{Baseline.}
We compare CARM primarily with Conformal Mask Calibration (CMC),
which calibrates a single threshold on the calibration split and
keeps it fixed throughout deployment. Within each
dataset--backbone combination, CMC and CARM receive exactly the same
frozen SR predictions and uncertainty maps and are evaluated on the
same chronologically ordered test stream. The sequential deployment
protocol and CARM configuration are also kept unchanged across
backbones. Therefore, each paired comparison between CMC and CARM
isolates the effect of online threshold adaptation rather than changes
in the underlying SR predictions or uncertainty estimates.

\textbf{Metrics.}
Let $T$ denote the number of test days. Using the realized
mask-conditioned risk $r_t(\tau_t)$ and trusted-area ratio
$c_t(\tau_t)$, we report
\begin{align*}
\mathrm{Risk}
&=
\frac{1}{T}\sum_{t=1}^{T} r_t(\tau_t), \quad
\mathrm{VR}
=
\frac{1}{T}\sum_{t=1}^{T}
\mathbf{1}\!\left\{r_t(\tau_t)>\alpha\right\},\\
\mathrm{MAER}
&=
\frac{1}{T-6}\sum_{t=1}^{T-6}
\mathbf{1}\!\left\{
\frac{1}{7}\sum_{j=t}^{t+6}r_j(\tau_j)>\alpha
\right\},\\
\mathrm{TS}
&=
\frac{1}{T}\sum_{t=1}^{T}c_t(\tau_t).
\end{align*}
Here, VR measures day-level violations, MAER measures exceedances
over stride-1 seven-day windows, and TS measures the average fraction
of valid pixels retained in the deployed trust masks. Lower values
are preferred for Risk, VR, and MAER, whereas higher TS indicates a
larger trusted region.

\subsection{Main Results}
\label{subsec:main_results}

\textbf{CARM improves deployment-time reliability under temporal
distribution shift.}
The CMC and CARM rows in Table~\ref{tab:main_scs_results} compare
static CMC and CARM on the South China Sea Chl-a stream using the
FNO backbone across three target levels,
$\alpha\in\{0.05,0.10,0.15\}$.

\begin{table*}[!t]
\centering
\caption{Comparison of static CMC, fixed-rate experts, and CARM on
the South China Sea Chl-a stream using the FNO backbone. Lower values
are preferred for Risk, VR, and MAER, and the best value within each
target level is bolded. TS reports retained coverage and is not
ranked.}
\label{tab:main_scs_results}
\resizebox{\textwidth}{!}{%
\begin{tabular}{l cccc cccc cccc}
\toprule
& \multicolumn{4}{c}{$\alpha=0.05$}
& \multicolumn{4}{c}{$\alpha=0.10$}
& \multicolumn{4}{c}{$\alpha=0.15$} \\
\cmidrule(lr){2-5}
\cmidrule(lr){6-9}
\cmidrule(lr){10-13}

Method
& Risk $\downarrow$
& VR (\%) $\downarrow$
& MAER (\%) $\downarrow$
& TS (\%)
& Risk $\downarrow$
& VR (\%) $\downarrow$
& MAER (\%) $\downarrow$
& TS (\%)
& Risk $\downarrow$
& VR (\%) $\downarrow$
& MAER (\%) $\downarrow$
& TS (\%) \\
\midrule

CMC
& 0.0513 & 47.31 & 56.11 & 26.03
& 0.0989 & 43.49 & 47.29 & 57.51
& 0.1496 & 43.40 & 46.95 & 94.78 \\

\midrule

Single expert ($\eta=0.001$)
& 0.0354 & 13.02 & \textbf{5.47} & 10.14
& 0.0746 & 10.42 & 1.74 & 42.32
& 0.1230 & 16.41 & 10.76 & 71.59 \\

Single expert ($\eta=0.005$)
& 0.0354 & 13.02 & \textbf{5.47} & 10.14
& 0.0733 & 9.81 & 1.65 & 41.36
& 0.1191 & 15.28 & 9.64 & 68.42 \\

Single expert ($\eta=0.020$)
& \textbf{0.0353} & 13.02 & \textbf{5.47} & 10.15
& 0.0728 & 9.90 & 1.04 & 41.04
& \textbf{0.1184} & 14.24 & 9.20 & 68.18 \\

Single expert ($\eta=0.100$)
& 0.0354 & 13.19 & \textbf{5.47} & 10.27
& \textbf{0.0722} & 9.46 & \textbf{0.95} & 40.89
& 0.1202 & \textbf{14.06} & \textbf{8.59} & 69.85 \\

\midrule

CARM
& 0.0354 & \textbf{12.93} & 5.50 & 10.19
& 0.0737 & \textbf{9.38} & 1.22 & 41.73
& 0.1211 & 14.76 & 9.69 & 70.47 \\

\bottomrule
\end{tabular}%
}
\end{table*}

CARM reduces the realized risk, day-level violations, and sustained
exceedances relative to static CMC at every level: at
$\alpha=0.05$, VR falls from $47.31\%$ to $12.93\%$ and MAER from
$56.11\%$ to $5.50\%$; at $\alpha=0.10$, VR falls from $43.49\%$ to
$9.38\%$ and MAER from $47.29\%$ to $1.22\%$; at $\alpha=0.15$, VR
falls from $43.40\%$ to $14.76\%$ and MAER from $46.95\%$ to
$9.69\%$. These improvements are accompanied by a smaller trusted
region, reflecting the expected reliability--coverage trade-off of
adaptive masking rather than a cost-free increase in retained area.

\begin{figure}[t]
    \centering
    \includegraphics[width=\columnwidth]
    {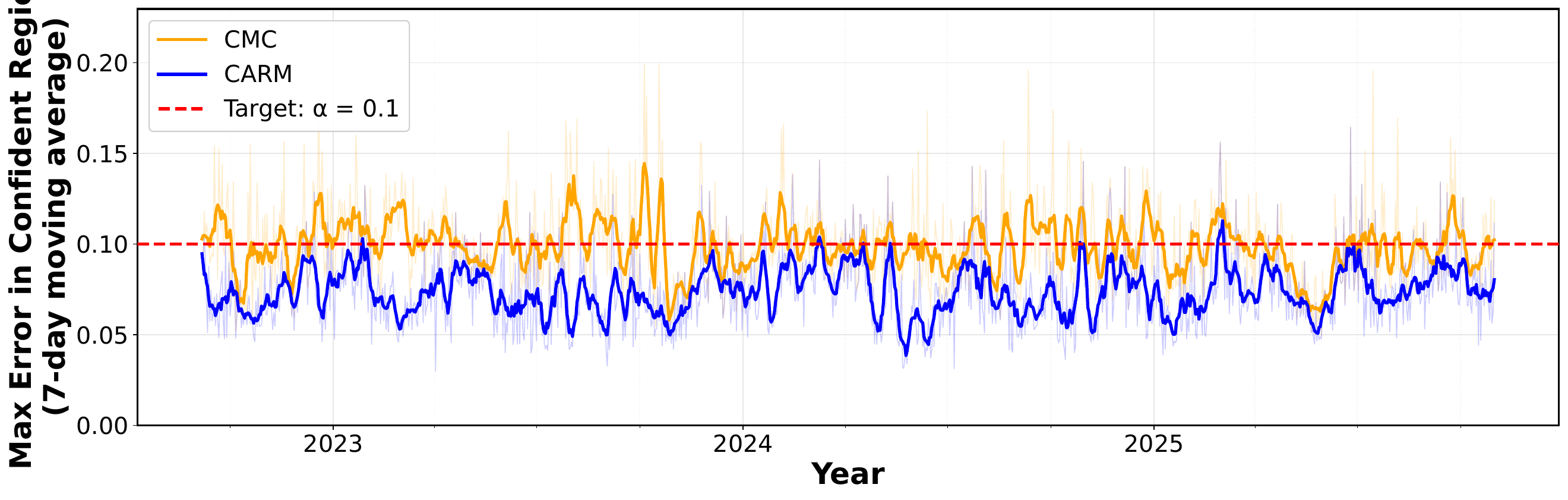}
    \caption{Temporal risk on the South China Sea stream at
    $\alpha=0.10$. The fixed CMC threshold exhibits sustained
    near-target or above-target periods under temporal distribution
    shift, whereas CARM adapts online using matured feedback. The
    dashed line denotes the target risk level $\alpha=0.10$.}
    \label{fig:motivation_static_misalignment}
\end{figure}

\textbf{Temporal behavior under distribution shift.}
Figure~\ref{fig:motivation_static_misalignment} provides an additional
temporal diagnostic on the South China Sea stream at $\alpha=0.10$.
The fixed CMC threshold exhibits sustained near-target or above-target
periods, whereas CARM adjusts its deployed threshold using matured
feedback. This comparison illustrates the deployment limitation of
one-shot calibration under temporal distribution shift.

\textbf{Effectiveness of CARM on multiple datasets and backbones.}
Table~\ref{tab:cross_domain} evaluates CARM at $\alpha=0.10$ across
three scientific datasets and multiple backbones.

\begin{table}[t]
\centering
\caption{Cross-domain and cross-backbone comparison at
$\alpha=0.10$.}
\label{tab:cross_domain}
\resizebox{\columnwidth}{!}{%
\begin{tabular}{llcccc}
\toprule
Dataset & Backbone & Method
& Risk $\downarrow$
& VR (\%) $\downarrow$
& MAER (\%) $\downarrow$ \\
\midrule

\multirow{12}{*}{Chl-a}
& \multirow{2}{*}{FNO}
& CMC  & 0.0989 & 43.49 & 47.29 \\
& & CARM & \textbf{0.0737} & \textbf{9.38} & \textbf{1.22} \\

& \multirow{2}{*}{SRNO}
& CMC  & 0.0974 & 41.24 & 38.38 \\
& & CARM & \textbf{0.0730} & \textbf{10.54} & \textbf{3.32} \\

& \multirow{2}{*}{HiNOTE}
& CMC  & 0.1014 & 43.08 & 43.29 \\
& & CARM & \textbf{0.0736} & \textbf{11.33} & \textbf{6.37} \\

& \multirow{2}{*}{MWT}
& CMC  & 0.0968 & 40.97 & 39.04 \\
& & CARM & \textbf{0.0655} & \textbf{11.07} & \textbf{2.39} \\

& \multirow{2}{*}{SwinIR}
& CMC  & 0.0981 & 41.37 & 43.16 \\
& & CARM & \textbf{0.0711} & \textbf{10.94} & \textbf{5.05} \\

& \multirow{2}{*}{EDSR}
& CMC  & 0.0977 & 39.92 & 44.22 \\
& & CARM & \textbf{0.0818} & \textbf{19.10} & \textbf{8.37} \\

\midrule

\multirow{6}{*}{ERA5}
& \multirow{2}{*}{FNO}
& CMC  & 0.0960 & 35.00 & 36.72 \\
& & CARM & \textbf{0.0844} & \textbf{9.72} & \textbf{6.21} \\

& \multirow{2}{*}{SRNO}
& CMC  & 0.0940 & 33.89 & 27.12 \\
& & CARM & \textbf{0.0768} & \textbf{10.28} & \textbf{3.67} \\

& \multirow{2}{*}{ResShift}
& CMC  & 0.0947 & 31.94 & 28.53 \\
& & CARM & \textbf{0.0782} & \textbf{11.11} & \textbf{3.67} \\

\midrule

\multirow{10}{*}{OSTIA}
& \multirow{2}{*}{MWT}
& CMC  & 0.2120 & 84.07 & 98.39 \\
& & CARM & \textbf{0.1513} & \textbf{64.73} & \textbf{68.54} \\

& \multirow{2}{*}{SRNO}
& CMC  & 0.1023 & 42.53 & 57.83 \\
& & CARM & \textbf{0.0677} & \textbf{11.27} & \textbf{5.15} \\

& \multirow{2}{*}{HiNOTE}
& CMC  & 0.1069 & 43.87 & 58.57 \\
& & CARM & \textbf{0.0703} & \textbf{10.80} & \textbf{6.16} \\

& \multirow{2}{*}{SwinIR}
& CMC  & 0.0980 & 39.07 & 52.14 \\
& & CARM & \textbf{0.0720} & \textbf{13.40} & \textbf{5.15} \\

& \multirow{2}{*}{ResShift}
& CMC  & 0.1052 & 40.60 & 51.94 \\
& & CARM & \textbf{0.0881} & \textbf{21.93} & \textbf{27.44} \\

\bottomrule
\end{tabular}%
}
\end{table}

The evaluated backbones include FNO~\cite{li2021fno},
MWT~\cite{gupta2021mwt}, SRNO~\cite{wei2023srno},
HiNOTE~\cite{luo2024hinote}, SwinIR~\cite{liang2021swinir},
EDSR~\cite{lim2017edsr}, and ResShift~\cite{yue2023resshift}.
CARM reduces all three primary deployment metrics over static CMC
across all fourteen combinations, confirming consistent relative
improvements regardless of domain or backbone choice. For example,
on OSTIA with the SRNO backbone, VR falls from $42.53\%$ to $11.27\%$
and MAER from $57.83\%$ to $5.15\%$. These results indicate that CARM
improves deployment-time reliability across diverse settings, even
though the magnitude of that improvement varies by dataset and
backbone.

\begin{figure*}[!t]
    \centering
    \includegraphics[width=0.8\textwidth]
    {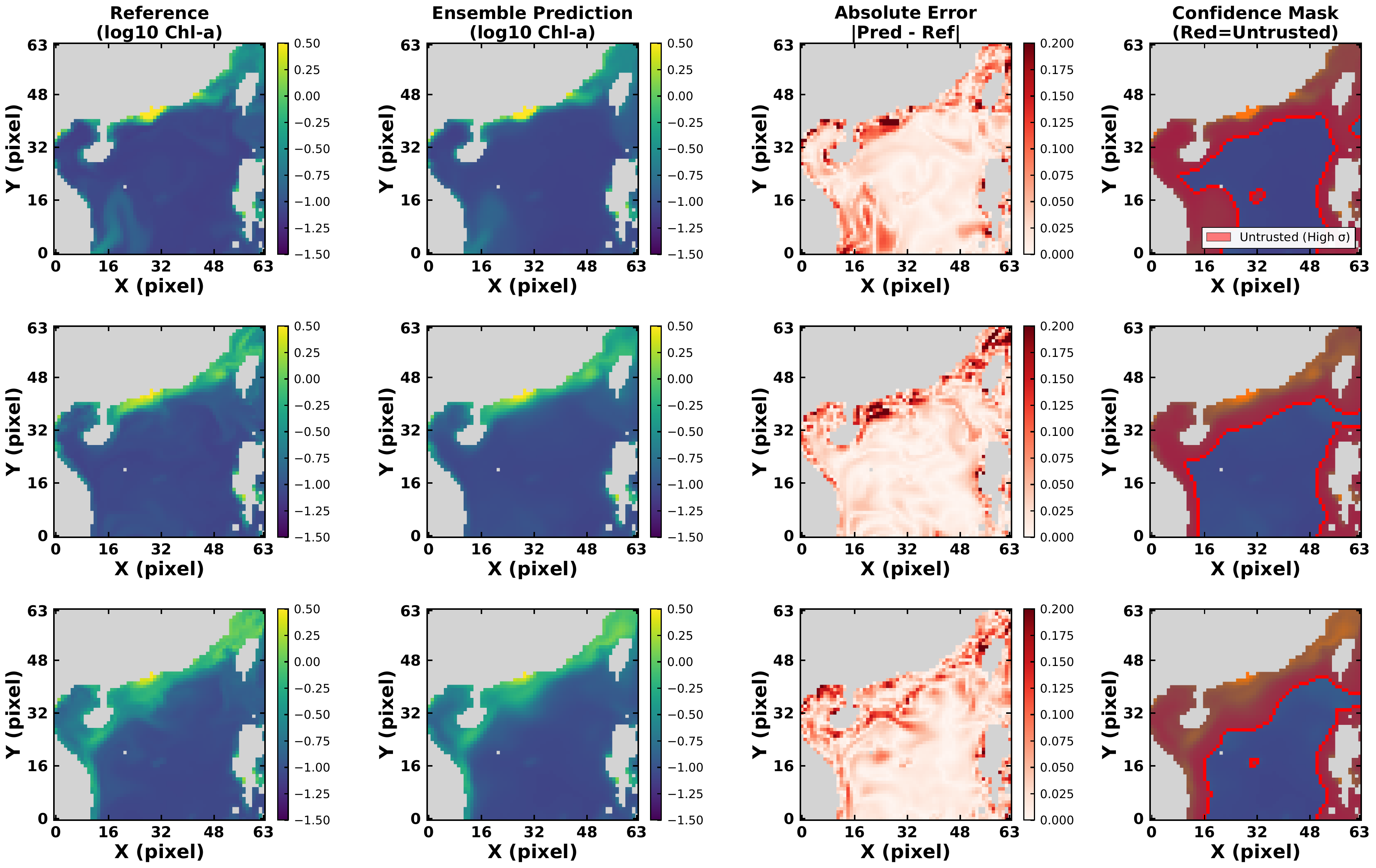}
    \caption{Qualitative examples of deployed trust masks on the South
    China Sea test stream at $\alpha=0.10$. Each row corresponds to
    one representative test sample. From left to right: HR reference,
    ensemble SR prediction, absolute reconstruction error, and the
    deployed trust mask. Red regions are rejected as untrusted.}
    \label{fig:qualitative_masks}
\end{figure*}

\textbf{Qualitative analysis of trust masks.}
Figure~\ref{fig:qualitative_masks} provides representative
qualitative examples on the South China Sea test stream at
$\alpha=0.10$. Regions with large reconstruction errors frequently
occur near complex coastal and boundary structures. The deployed
trust masks reject many of these spatially unreliable regions while
retaining more stable interior-ocean areas, illustrating the spatial
correspondence between predictive uncertainty and realized
reconstruction error.

\subsection{Ablation Studies}
\label{subsec:ablation}

\textbf{Comparison with fixed-rate experts.}
The fixed-rate expert rows in
Table~\ref{tab:main_scs_results} compare CARM with four
single-expert controllers using fixed adaptation rates
$\eta\in\{0.001,0.005,0.02,0.1\}$.

The lowest-risk fixed rate is not stable across $\alpha$:
$\eta=0.02$ attains the lowest Risk at $\alpha=0.05$ and
$\alpha=0.15$, whereas $\eta=0.1$ attains the lowest Risk at
$\alpha=0.10$, so committing to any single rate in advance would
sacrifice performance at some target levels. CARM instead aggregates
across rates online, attaining the lowest VR among all methods at
both $\alpha=0.05$ and $\alpha=0.10$. At $\alpha=0.10$
specifically, $\eta=0.1$ attains the lowest Risk and MAER among the
fixed-rate experts, yet CARM still retains a larger trusted area
($\mathrm{TS}=41.73\%$ vs.\ $40.89\%$). CARM's improved risk control
is therefore not an artifact of aggressively shrinking the trusted
region, but reflects adaptive threshold calibration that responds to
the evolving data distribution.

\textbf{Effect of expert-pool composition.}
We evaluate how the composition of the expert pool affects online
risk control on the South China Sea stream at $\alpha=0.10$.
Table~\ref{tab:expert_pool} compares several subsets of adaptation
rates under the same delayed-feedback protocol. Pools containing
faster experts generally achieve lower average risk and sustained
exceedance rates, whereas the full heterogeneous pool achieves the
lowest empirical violation rate. No pool configuration dominates all
reported metrics, supporting the use of heterogeneous response rates
rather than a single prescribed subset.

\begin{table}[t]
\centering
\caption{Effect of expert-pool composition on the South China Sea
stream at $\alpha=0.10$ using the FNO backbone. Lower is better.}
\label{tab:expert_pool}
\resizebox{\columnwidth}{!}{%
\begin{tabular}{lccc}
\toprule
Expert Pool & Risk $\downarrow$ & VR $\downarrow$
& MAER $\downarrow$ \\
\midrule

$\{0.001,0.005\}$
& 0.0742 & 9.90\% & 1.75\% \\

$\{0.02,0.1\}$
& \textbf{0.0727} & 9.55\% & \textbf{1.05\%} \\

$\{0.001,0.02,0.1\}$
& 0.0735 & 9.64\% & \textbf{1.05\%} \\

$\{0.001,0.005,0.02,0.1\}$
& 0.0737 & \textbf{9.38\%} & 1.22\% \\

\bottomrule
\end{tabular}%
}
\end{table}

\textbf{Sensitivity to the matured window length.}
Table~\ref{tab:window_length_all_alphas} evaluates the matured window
length $w\in\{30,60,90\}$ across all three target risk levels,
$\alpha\in\{0.05,0.10,0.15\}$, on the South China Sea test stream.
The results show the expected responsiveness--stability trade-off:
short windows are more sensitive to transient fluctuations, whereas
long windows provide stronger smoothing but may adapt more slowly.
Overall, $w=60$ provides a stable compromise across the evaluated
target levels, although it is not optimal for every metric.

\begin{table}[!t]
\centering
\caption{Sensitivity to the matured window length on the South China
Sea test stream. VR and MAER are reported in percent. Lower is
better.}
\label{tab:window_length_all_alphas}
\resizebox{\columnwidth}{!}{%
\begin{tabular}{ccccc}
\toprule
$\alpha$ & $w$
& Risk $\downarrow$
& VR $\downarrow$
& MAER $\downarrow$ \\
\midrule

\multirow{3}{*}{$0.05$}
& 30 & 0.0366 & 14.76\% & 6.28\% \\
& 60 & \textbf{0.0354} & \textbf{12.93\%} & 5.50\% \\
& 90 & 0.0355 & 13.19\% & \textbf{5.15\%} \\

\midrule

\multirow{3}{*}{$0.10$}
& 30 & 0.0738 & 10.68\% & 1.92\% \\
& 60 & 0.0737 & \textbf{9.38\%} & \textbf{1.22\%} \\
& 90 & \textbf{0.0732} & 9.46\% & 1.31\% \\

\midrule

\multirow{3}{*}{$0.15$}
& 30 & 0.1231 & 15.80\% & 10.03\% \\
& 60 & \textbf{0.1211} & \textbf{14.76\%}
& \textbf{9.69\%} \\
& 90 & 0.1214 & 15.54\% & 11.17\% \\

\bottomrule
\end{tabular}%
}
\end{table}

\section{Conclusion}
We investigated reliable scientific super-resolution under temporal distribution shift, where spatial trust masks calibrated once on historical data can lose reliability as the deployment distribution evolves. To address this problem, we proposed CARM, a model-agnostic post-processing framework that causally adapts trust-mask thresholds from matured feedback through multi-rate expert aggregation, without retraining or modifying the underlying super-resolution model. Experiments across multiple geophysical datasets and diverse super-resolution backbones show that CARM consistently reduces average risk, violation rate, and mean accumulated excess risk relative to static CMC, although the magnitude of the improvement depends on the dataset and backbone. These results indicate that the benefit of CARM arises from better aligning the trusted region with the evolving error distribution rather than from improving the underlying reconstruction itself. This reliability is not cost-free: stricter risk control can reduce trusted spatial coverage, and effective adaptation depends on the uncertainty scores providing an informative ranking of spatial reconstruction errors. Moreover, CARM is designed to control risk over time rather than eliminate every instantaneous violation, while delayed feedback inevitably limits its response to abrupt distribution changes. Future work should therefore improve uncertainty ranking and calibration, develop faster adaptation mechanisms for sudden shifts, and incorporate explicit risk--coverage objectives to preserve sufficient usable regions while maintaining reliable deployment.

\clearpage

\balance
\bibliographystyle{IEEEtran}
\bibliography{references}

\clearpage

\appendices

\section{Proof of Theorem~\ref{thm:main}}
\label{app:proof}

Write $\tau_{\min}:=\min\mathcal{G}$ and
$\eta_{\min}:=\min_k\eta_k$.

\begin{assumption}
\label{asm:all}
\emph{(A1)} $\tilde e_t(p)\in[0,1]$ for all $t,p$, hence
$r_t(\tau)\in[0,1]$ by \eqref{eq:risk}.\\
\emph{(A2)} $\tau\le\tau'\Rightarrow
M_t(\tau)\subseteq M_t(\tau')$ by \eqref{eq:trust_mask}, so
$r_t(\cdot)$ is nondecreasing.\\
\emph{(A3)} $\tau_t$ is $\mathcal F_t$-measurable and $r_t$ matures
at $t+d$ with $d<\infty$, where
$\mathcal F_t=\sigma(X_{1:t},\hat Y_{1:t},U_{1:t},Y_{1:t-d})$.\\
\emph{(A4)} Let
$\mathcal S_t:=\arg\min_{\tau\in\mathcal G_t}\hat q_t(\tau)$ and
$\tau^{\mathrm{fb}}_t:=\max\mathcal S_t$, the candidate returned by
\eqref{eq:expert_threshold_selection} when the internal targets are
nonpositive.
\end{assumption}

Assumptions (A1) and (A2) hold by construction and (A3) by the
deployment protocol; (A4) is the only substantive restriction. Three
features of
\eqref{eq:expert_adaptation}--\eqref{eq:expert_threshold_selection}
are load-bearing: the internal states $\alpha^{e,(k)}_t$ are
unprojected, the driving signal $\bar v_t$ is shared across experts
and computed from the deployed decisions, and ties break downward.
Expert heterogeneity is carried entirely by the rates $\eta_k$ and
the weights $\omega^{(k)}_t$; the candidate-side window $w$ of
$\hat q_t(\cdot)$ plays no role in the constant below.

\begin{proof}
Throughout write
$n:=w+d$, $H_k:=\alpha'/\eta_k$,
\[
\pi^{(k)}_T
:=
\frac{1}{T}
\sum_{t\le T}
\mathbf 1\{r_t(\tau^{(k)}_t)>\alpha'\},
\]
and
\[
\pi_T
:=
\frac{1}{T}
\sum_{t\le T}
\mathbf 1\{r_t(\tau_t)>\alpha'\}.
\]

\emph{Step 1 (per-expert linearization).}
Set
\[
G^{(k)}_t
:=
\sum_{s=1}^{t-1}
\left(\bar v^{(k)}_s-\alpha'\right),
\qquad
G^{(k)}_1=0.
\]
By (A5) and induction,
\eqref{eq:expert_adaptation} gives
\begin{align*}
\alpha^{e,(k)}_{t+1}
&=
\alpha'
-\eta_kG^{(k)}_t
+\eta_k\left(\alpha'-\bar v^{(k)}_t\right)\\
&=
\alpha'-\eta_kG^{(k)}_{t+1},
\end{align*}
hence
$\alpha^{e,(k)}_t=\alpha'-\eta_kG^{(k)}_t$.
The absence of a projection is what makes each recursion exactly
linear.

\emph{Step 2 (saturation).}
Let $H:=\alpha'/\eta_{\min}$ and suppose $G_t>H$. Then $G_t>0$ and
$\eta_kG_t\ge\eta_{\min}G_t>\alpha'$, so
$\alpha^{e,(k)}_t<0$ for every $k$ by Step 1. Since
$\hat q_t\ge0$, the objective in
\eqref{eq:expert_threshold_selection} equals
$\hat q_t(\tau)-\alpha^{e,(k)}_t$, which differs across experts only
by a constant independent of $\tau$; minimizing it is therefore
equivalent to minimizing $\hat q_t$, and all $K$ experts share the
minimizer set $\mathcal S_t$. By (A2), each $r_j(\cdot)$ is
nondecreasing, hence so is
$\tau\mapsto\mathbf 1\{r_j(\tau)>\alpha'\}$ and therefore
$\hat q_t(\cdot)$. Its minimum over $\mathcal G_t$ is attained on a
plateau, and the tie-breaking rule of
\eqref{eq:expert_threshold_selection} selects its right endpoint
$\tau^{\mathrm{fb}}_t$ for every $k$. All $K$ proposals coincide, so
$\tau_t=\tau^{\mathrm{fb}}_t$ for any weights, and
$r_t(\tau_t)\le\alpha'$ by (A4).

\emph{Step 2 (per-expert saturation).}
Suppose $G^{(k)}_t>H_k$. Then $\alpha^{e,(k)}_t<0$, and since
$\hat q_t\ge0$, the objective in
\eqref{eq:expert_threshold_selection} equals
$\hat q_t(\tau)-\alpha^{e,(k)}_t$, so expert $k$ minimizes
$\hat q_t$. By (A2), each $r_j(\cdot)$ is nondecreasing, hence so is
$\hat q_t(\cdot)$; its minimizer set is a plateau and the
tie-breaking rule returns its right endpoint. Thus
$\tau^{(k)}_t=\tau^{\mathrm{fb}}_t$ and
$r_t(\tau^{(k)}_t)\le\alpha'$ by (A4). Experts need not saturate
simultaneously; the argument is applied to each $k$ separately.

\emph{Step 3 (delayed reflection).}
Fix $k$; we claim
\[
G^{(k)}_t
\le
G^{(k)}_{\max}
:=
H_k+n(1-\alpha').
\]
Increments obey
\[
G^{(k)}_{t+1}-G^{(k)}_t
\in[-\alpha',1-\alpha'].
\]
Let $T_0$ be minimal with
$G^{(k)}_{T_0}>G^{(k)}_{\max}$.
If $T_0\le n$, then
\[
G^{(k)}_{T_0}
\le
(n-1)(1-\alpha')
<
G^{(k)}_{\max}.
\]
Otherwise, for $0\le j\le n$,
\[
G^{(k)}_{T_0-j}
\ge
G^{(k)}_{T_0}-j(1-\alpha')
>
H_k,
\]
so expert $k$ is saturated on every
$s\in\{T_0-n,\dots,T_0-1\}$ and, by Step 2, none of these days is a
violation of its own proposal. Its buffer at $T_0-1$ is
$\{T_0-n,\dots,T_0-1-d\}$, so
$\bar v^{(k)}_{T_0-1}=0$ and
$G^{(k)}_{T_0}<G^{(k)}_{T_0-1}$, a contradiction.

\emph{Step 4 (per-expert conversion).}
Exchanging summation as before,
\[
\sum_{t\le T}\bar v^{(k)}_t
=
\sum_j
\kappa_j
\mathbf 1
\{r_j(\tau^{(k)}_j)>\alpha'\},
\]
with $\kappa_j\in[0,1]$ and $\kappa_j<1$ for at most $n-1$ days, so
\[
T\pi^{(k)}_T
\le
T\alpha'+C^{(k)},
\]
where
\[
C^{(k)}
:=
H_k+n(2-\alpha')-1.
\]

\emph{Step 5 (aggregation).}
Since $\omega_t$ lies in the simplex,
$\tau_t\le\max_k\tau^{(k)}_t$, so
$r_t(\tau_t)\le\max_k r_t(\tau^{(k)}_t)$ by (A2), and
\[
\mathbf 1\{r_t(\tau_t)>\alpha'\}
\le
\sum_k
\mathbf 1\{r_t(\tau^{(k)}_t)>\alpha'\}.
\]
Hence
\[
\pi_T
\le
\sum_k\pi^{(k)}_T
\le
K\alpha'
+
\frac{\bar C}{T},
\]
with
\[
\bar C
:=
\sum_k C^{(k)}
=
\alpha'
\sum_k\eta_k^{-1}
+
K\bigl[n(2-\alpha')-1\bigr].
\]
This step uses no property of the weights beyond the simplex
constraint.

By (A1), $r_t(\tau_t)\le\alpha'$ on non-violation days and
$r_t(\tau_t)\le1$ otherwise, so
\begin{align*}
\frac{1}{T}\sum_{t=1}^{T}r_t(\tau_t)
&\le
\alpha'+\pi_T(1-\alpha')\\
&\le
\alpha'\bigl[1+K(1-\alpha')\bigr]
+
\frac{(1-\alpha')\bar C}{T}.
\end{align*}
Let $\alpha'$ be the smaller root of
\[
K\alpha'^2-(K+1)\alpha'+\alpha=0,
\]
so that the leading term equals $\alpha$. Then
\begin{align*}
\frac{1}{T}\sum_{t=1}^{T}r_t(\tau_t)
&\le
\alpha
+
\frac{1-\alpha'}{T}
\left[
\alpha'\sum_{k=1}^{K}\frac{1}{\eta_k}
+
K\bigl(n(2-\alpha')-1\bigr)
\right]\\
&=
\alpha+O(1/T)
\end{align*}
for $T\ge1$. The bracket is independent of $T$, giving
\eqref{eq:main} at an $O(1/T)$ rate.
\end{proof}

\end{document}